\documentclass{ijclclp}
\usepackage{float}

\title{Diffusion Models for conditional MRI generation}

\author{
    Miguel Herencia García del Castillo\textsuperscript{1},  \\
    Ricardo Moya Garcia\textsuperscript{2}, \\
    Manuel Jesús Cerezo Mazón\textsuperscript{1},  \\
    Ekaitz Arriola Garcia\textsuperscript{1},  \\
    Pablo Menéndez Fernández-Miranda\textsuperscript{1} \\
    \textsuperscript{1}Ainovis.health
    }
    \affiliation{\textsuperscript{1} Ainovis, \textsuperscript{2} Telefónica Innovación Digital}  
    \email{\textsuperscript{1} \texttt{info@ainovis.health}
}

\begin{document}

\maketitle
\thispagestyle{firstpage}

\begin{abstract}
In this article, we present a Latent Diffusion Model (LDM) for the generation of brain Magnetic Resonance Imaging (MRI), conditioning its generation based on pathology (Healthy, Glioblastoma, Sclerosis, Dementia) and acquisition modality (T1w, T1ce, T2w, Flair, PD).

To evaluate the quality of the generated images, the Fréchet Inception Distance (FID) and Multi-Scale Structural Similarity Index (MS-SSIM) metrics were employed. The results indicate that the model generates images with a distribution similar to real ones, maintaining a balance between visual fidelity and diversity. Additionally, the model demonstrates extrapolation capability, enabling the generation of configurations that were not present in the training data.

The results validate the potential of the model to increase in the number of samples in clinical datasets, balancing underrepresented classes, and evaluating AI models in medicine, contributing to the development of diagnostic tools in radiology without compromising patient privacy.

\end{abstract}

\vspace{0.5cm}
\section{Introduction}

Artificial intelligence (AI) has transformed medical image analysis, enabling the automation of diagnostic and segmentation tasks with unprecedented accuracy \cite{bakas2018brats, ho2020denoising}. However, the development of AI models in this field faces significant challenges due to the scarcity of clinical data, strict privacy regulations, and the high costs associated with annotation and labeling \cite{whang2023, panhuis2014, malin2018}. The collection of large volumes of medical data is a complex and costly process, constrained by the difficulty of accessing representative images of various pathologies and MRI modalities. Additionally, regulations such as the General Data Protection Regulation (GDPR) in Europe and the American Data Privacy and Protection Act (ADPPA) in the United States restrict the use and sharing of medical data, limiting access to extensive and diverse databases.

To address these challenges, medical image generation has emerged as a key strategy. By creating artificial data that mimic the characteristics of real images, generative models enable dataset augmentation without compromising patient privacy, facilitate the evaluation of AI models, and improve the representation of rare pathologies \cite{rombach2022, pinaya2022}. In this context, multiple approaches to medical image generation have been developed, with Generative Adversarial Networks (GANs) \cite{goodfellow2014gan} and Diffusion Models (DMs) \cite{dhariwal2021diffusion} standing out. While these models have proven effective in image generation, they present limitations such as high computational cost, dependence on training data, and a lack of generalization and diversity in generated images \cite{zaman2024}.

In this work, we propose a model based on Latent Diffusion Models (LDMs) for conditional MRI generation, allowing the specification of both pathology and acquisition modality. This approach optimizes the image generation process by performing diffusion in a compressed latent space, reducing computational load without compromising visual quality. The model is capable of generating images in any combination of pathology and modality, even in configurations absent from the training data, demonstrating its extrapolation capability and potential to improve the diversity of medical datasets \cite{guo2024maisi}.

Conditioned synthetic medical image generation offers multiple advantages. First, it increases the diversity of clinical datasets, facilitating the training of AI models with a broader representation of clinical cases. Additionally, it helps mitigate data imbalance by generating images for pathologies and modalities that are underrepresented in real data \cite{fernandez2024}. It also facilitates the validation of diagnostic models by creating synthetic data for controlled testing and, finally, ensures patient privacy, as the generated data do not contain identifiable information.

The remainder of this article is organized as follows. Section 2 describes the proposed model for generating conditioned synthetic medical images, detailing its architecture and workflow. Section 3 presents the experiments and results, including the evaluation of image quality using quantitative metrics and visual examples. Finally, Section 4 discusses the conclusions and future research directions, highlighting the model’s significance and potential applications in AI for medicine.

\vspace{0.5cm}
\section{Solution}
\subsection{Generative AI Models in Imaging}
The generation of medical images using artificial intelligence has evolved significantly, giving rise to multiple approaches that have demonstrated their effectiveness in this field. Among the most commonly used models are: Generative Adversarial Networks (GANs), Variational Autoencoders (VAEs) \cite{kingma2013vae}, Flow-Based Models \cite{dinh2016density}, and Diffusion Models (DMs), each with specific characteristics that make them suitable for different applications.

A particularly efficient approach is the Latent Diffusion Model (LDM), which optimizes image generation by performing diffusion in a compressed latent space. This approach reduces computational load without sacrificing visual quality, making it an ideal option for the synthesis of medical images conditioned by pathology and modality.

\vspace{1cm}
The main characteristics of these approaches are presented below:

\begin{itemize}
    \item \textbf{GANs}: They use two competing neural networks in an adversarial framework: a generator, which creates synthetic images, and a discriminator, which attempts to distinguish between real and generated images \cite{goodfellow2014gan}. Thanks to this approach, GANs have been widely used in medical image synthesis and resolution enhancement \cite{kazeminia2020gans}. However, they suffer from mode collapse, where the generator produces only a subset of images, and training instability, which can prevent the model from converging \cite{salimans2016improved}.

    \item \textbf{VAEs}: They model the latent distribution of data, enabling the generation of new images by sampling from that distribution. They consist of an encoder, which compresses the image into a latent space, and a decoder, which reconstructs it. Their main advantage is stability in training, as they do not rely on an adversarial strategy \cite{doersch2016tutorial}. However, the generated images tend to be less sharp compared to those produced by GANs, as they optimize reconstruction in probabilistic terms \cite{razavi2019vqvae}.

    \item \textbf{Flow-Based Models}: They apply invertible transformations to model the probability of the data, allowing precise control over the attributes of the generated images. Unlike other models, they can calculate the exact probability of each sample, facilitating its interpretation \cite{kingma2018glow}. They do not suffer from mode collapse, but their high computational cost and complex architecture make them less efficient for training and image generation \cite{papamakarios2019normalizing}.

    \item \textbf{Diffusion Models (DMs)}: These models generate images by iteratively removing noise from a random distribution, gradually denoising a sample towards a realistic image \cite{song2021score}. They are highly effective in the synthesis of high-quality images and do not suffer from mode collapse issues. However, their main limitation is their high computational cost, as multiple inference steps are required to generate an image. 
    
    The forward diffusion process is described as:
    \begin{equation}
        q(x_t | x_{t-1}) = \mathcal{N}(x_t; \sqrt{\alpha_t} x_{t-1}, (1 - \alpha_t) I)
    \end{equation}
    where $x_t$ represents the image at time step $t$, $\alpha_t$ controls the noise schedule, and $I$ is the identity matrix.
    
    The reverse process learns to denoise step by step:
    \begin{equation}
        p(x_{t-1} | x_t) = \mathcal{N}(x_{t-1}; \mu_\theta(x_t, t), \Sigma_\theta(x_t, t))
    \end{equation}
    where $\mu_\theta$ and $\Sigma_\theta$ are the predicted mean and variance of the denoised image at each step.

\end{itemize}

\textbf{Latent Diffusion Models (LDMs)} combine diffusion models with autoencoders, allowing them to operate in a compressed latent space instead of complete images. This optimizes the image synthesis process, reducing computational cost without compromising visual quality. In this work, we explore the role of LDMs in the generation of conditioned medical images, highlighting their ability to adapt to multiple pathologies and modalities.

\vspace{0.5cm}
\section{Proposed Solution}
The model proposed in this work is a \textit{Latent Diffusion Model} (LDM) designed to generate conditioned magnetic resonance imaging (MRI). As analyzed in the experiments and results section, image generation is performed through a conditioning scheme, allowing specification of both pathology ({Healthy, Glioblastoma, Sclerosis, Dementia}) and MRI modality (\textit{T1w, T1ce, T2w, Flair, PD}) \cite{lupke2024physics}.

This approach enables the generation of medical images with precise control over their characteristics, facilitating the creation of synthetic MRI scans in specific configurations. For example, the model can generate an image of a healthy brain in \textit{T1w} modality or an MRI of a brain with multiple sclerosis in \textit{FLAIR} modality, ensuring that the anatomical structure and appearance are consistent with the defined parameters.

Since the model's implementation is part of a company's intellectual property, internal details about its operation and specific training process will not be disclosed. However, a general description of its architecture and workflow is provided, explaining the key components in the training and inference stages.

The proposed model is based on a latent diffusion architecture, where image generation occurs in a compressed latent space instead of operating directly on the original image. This allows for efficient reconstruction with lower computational cost.

The system consists of three main modules:

\begin{itemize}
    \item \textbf{Encoder}: Converts MRI images into a compact latent representation, reducing dimensionality and facilitating model processing. The encoder maps an input image $x$ to a latent space $z$, parameterized by a mean $\mu(x)$ and a variance $\sigma^2(x)$:
    
    \begin{equation}
        q(z | x) = \mathcal{N}(z; \mu(x), \sigma^2(x))
    \end{equation}
    
    \item \textbf{U-Net network with DDPM}: During the training process of a diffusion model, we start with a clean latent representation \( z_0 \). Gaussian noise is progressively added following a predefined schedule to generate a noisy latent \( z_t \) at each timestep \( t \): \[ z_t = \sqrt{\bar{\alpha}_t}\, z_0 + \sqrt{1-\bar{\alpha}_t}\, \epsilon, \] where \( \epsilon \sim \mathcal{N}(0,I) \) is Gaussian noise and \( \bar{\alpha}_t \) is the cumulative product of the noise schedule coefficients. The conditional U-Net receives the noisy latent \( z_t \), the timestep \( t \), and additional conditioning information (such as pathology or modality) to predict the noise that was added, denoted as \( \epsilon_\theta(z_t, t) \). The training objective is to minimize error between the true noise \( \epsilon \) and the U-Net's prediction: \[\mathcal{L} = \mathbb{E}_{z_0,\epsilon,t}\left[\|\epsilon - \epsilon_\theta(z_t, t)\|^2\right]. \] By optimizing this loss, the U-Net learns to accurately predict the noise residual. \cite{DBLP:journals/corr/abs-2006-11239}
    
    \item \textbf{Decoder}: Reconstructs the final synthetic image from the refined latent space, generating an MRI with the specified characteristics. The decoder maps the latent vector $z$ back to the image space $\hat{x}$ through a learned transformation $p(x | z)$:
    
    \begin{equation}
        \hat{x} = g_{\theta}(z)
    \end{equation}
    
    where $g_{\theta}$ is the decoder function parameterized by a neural network.
\end{itemize}

\vspace{0.3cm}
The use of compressed latent spaces improves computational efficiency without compromising the visual quality of the generated images.The model is trained using a diffusion scheme in the latent space, allowing the neural network to learn how to reconstruct synthetic images from a degraded representation.

\vspace{0.3cm}

\begin{figure}[H]
    \centering
    \includegraphics[width=0.85\textwidth]{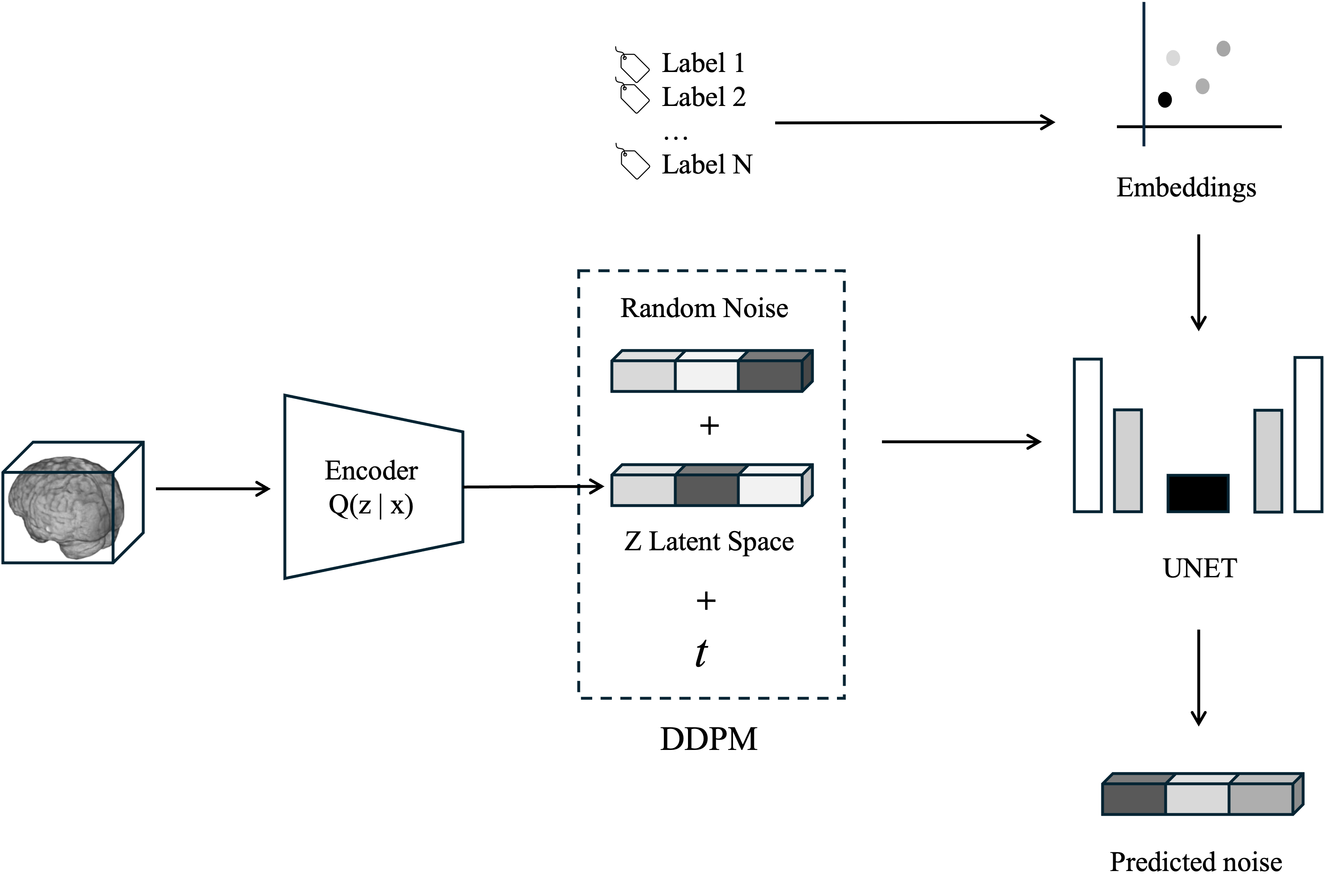} 
    \caption{Training process scheme of the generative model.}
    \label{fig:entrenamiento}
\end{figure}

This scheme, shown in Figure~\ref{fig:entrenamiento}, allows the model to learn anatomical patterns from the training data, ensuring that the generated images are realistic and representative of various clinical configurations.

\begin{itemize}
    \item \textbf{Inference with DDIM}: Once trained, the model can generate new synthetic images without requiring a real input image. During inference, a simplified process is followed compared to training:
    
    \begin{enumerate}
        \item \textbf{Random latent space initialization}: A random starting point is generated in the latent space.
        \item \textbf{Conditioning}: The corresponding \textit{embedding} for the desired pathology and modality is introduced.
        \item \textbf{DDIM Sampling}: The deterministic DDIM (Denoising Diffusion Implicit Model) method is applied, refining the noise latent space by iteratively predicting and removing noise:
        \begin{equation}
            z_{t-1} = \sqrt{\alpha_{t-1}} \left( \frac{z_t - \sqrt{1 - \alpha_t} \epsilon_\theta (z_t, t)}{\sqrt{\alpha_t}} \right) + \sigma_t \epsilon
        \end{equation}
        where $z_t$ is the latent variable at time step $t$, $\alpha_t$ controls the noise schedule, and $\sigma_t$ is the noise scaling factor. \cite{song2022denoisingdiffusionimplicitmodels}
        \item \textbf{Reconstruction with the decoder}: The refined latent representation is converted into a final image in the original MRI domain.
    \end{enumerate}
\end{itemize}

\begin{figure}[H]
    \centering
    \includegraphics[width=0.85\textwidth]{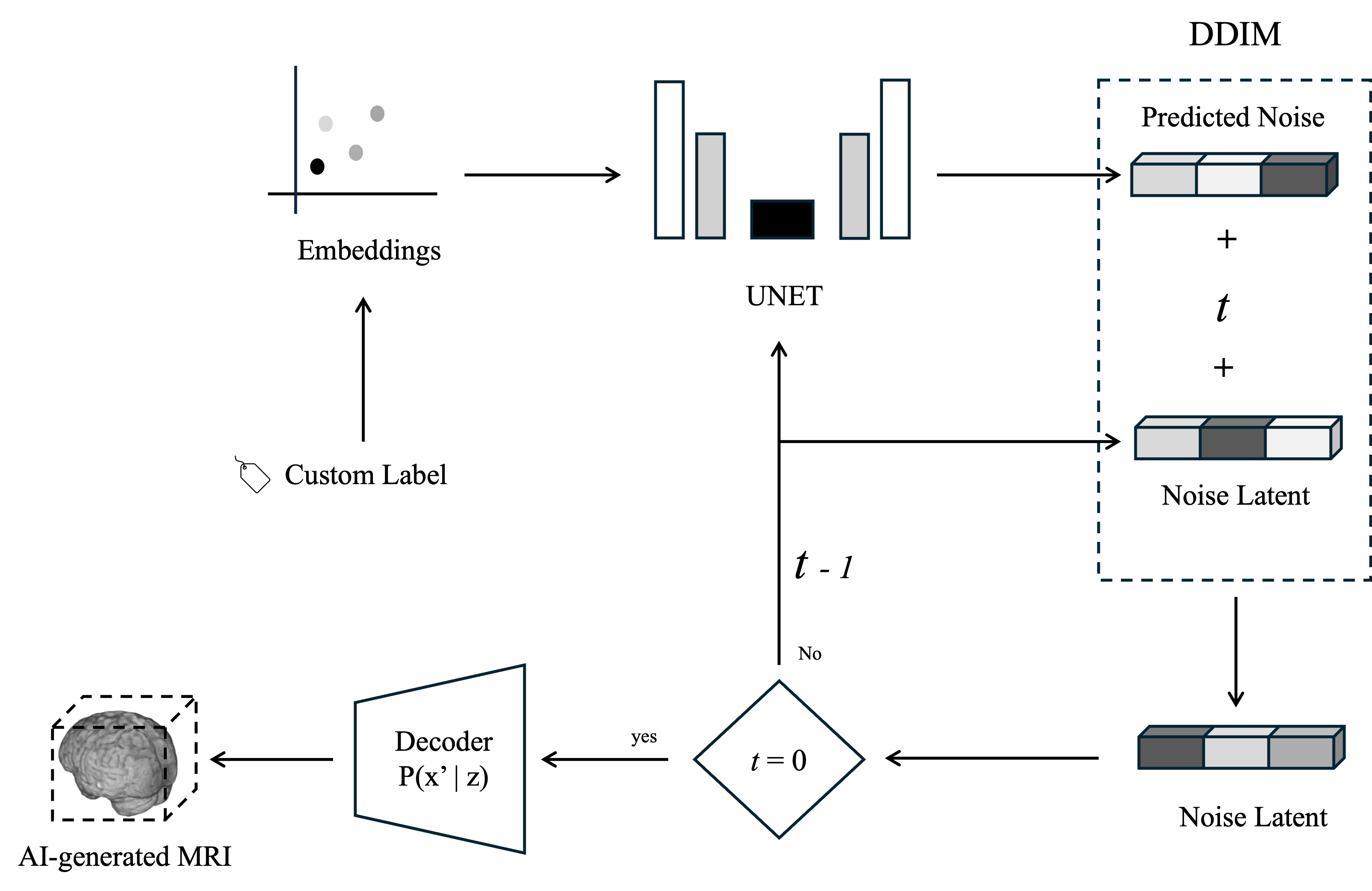}
    \caption{Inference process scheme of the generative model.}
    \label{fig:inferencia}
\end{figure}

This process, illustrated in Figure~\ref{fig:inferencia}, enables the generation of high-quality synthetic images, maintaining realistic anatomical structures and ensuring that the conditioning specifications are accurately reflected.

\section{Experiments and Results}
\subsection{Datasets}
To evaluate the performance of the conditioned medical image generator model, various brain MRI datasets were employed, representing different pathologies and acquisition modalities. The selection of these datasets aimed to ensure a robust evaluation of the model under multiple clinical conditions.
\vspace{0.3cm}
The data used come from four main sources: IXI and OASIS (for healthy subjects and dementia), BRATS2021 (glioblastoma), and ISBI 2015 Challenge along with open\_ms\_data (multiple sclerosis). Table~\ref{tab:datasets} summarizes the characteristics of each dataset, including the represented pathology and available modalities.
\vspace{0.3cm}
\begin{table}[h]
    \centering
    \renewcommand{\arraystretch}{1.2} 
    \begin{tabular}{|l|l|l|}
        \hline
        \textbf{Pathology} & \textbf{Dataset} & \textbf{Modality} \\
        \hline
        \textbf{Healthy} & IXI, OASIS-1, OASIS-2 & T1w, T2w, PD \\
        \textbf{Glioblastoma} & BRATS2021 & T1w, T1ce, T2w, Flair \\
        \textbf{Sclerosis} & ISBI 2015 Challenge, open\_ms\_data & T1w, T1ce, T2w, Flair, PD \\
        \textbf{Dementia} & OASIS-1, OASIS-2 & T1w \\
        \hline
    \end{tabular}
    \caption{Datasets used for model evaluation.}
    \label{tab:datasets}
\end{table}
\vspace{0.3cm}
Table~\ref{tab:distribucion_imagenes} details the distribution of images by pathology and MRI modality, while Table~\ref{tab:porcentaje_imagenes} presents the percentage representation of those images within the dataset.

\begin{table}[h]
    \centering
    \begin{tabular}{|l|c|c|c|c|c|}
        \hline
        \textbf{Pathology} & \textbf{T1w} & \textbf{T1ce} & \textbf{T2w} & \textbf{FLAIR} & \textbf{PD} \\
        \hline
        Healthy & 957 & - & 578 & - & 578 \\
        Glioblastoma & 887 & 887 & 887 & 887 & - \\
        Sclerosis & 49 & 30 & 49 & 49 & 19 \\
        Dementia & 139 & - & - & - & - \\
        \hline
    \end{tabular}
    \caption{Distribution of images by pathology and modality.}
    \label{tab:distribucion_imagenes}
\end{table}

\begin{table}[h]
    \centering
    \begin{tabular}{|l|c|c|c|c|c|}
        \hline
        \textbf{Pathology} & \textbf{T1w} & \textbf{T1ce} & \textbf{T2w} & \textbf{FLAIR} & \textbf{PD} \\
        \hline
        Healthy & 15.9\% & - & 9.6\% & - & 9.6\% \\
        Glioblastoma & 14.7\% & 14.7\% & 14.7\% & 14.7\% & - \\
        Sclerosis & 0.8\% & 0.5\% & 0.8\% & 0.8\% & 0.3\% \\
        Dementia & 2.3\% & - & - & - & - \\
        \hline
    \end{tabular}
    \caption{Percentage distribution of images by pathology and modality.}
    \label{tab:porcentaje_imagenes}
\end{table}

In total, the dataset used consists of 5,996 MRI scans, distributed among different pathologies and modalities. However, an imbalance in image distribution is observed, as some pathologies have greater representation than others. In particular, there is a higher number of images available for healthy subjects and glioblastoma patients, whereas the samples corresponding to multiple sclerosis and dementia are significantly lower.

This imbalance in data distribution may affect the model's evaluation in underrepresented configurations. However, the proposed model has demonstrated the ability to generate synthetic images in combinations not present in the training dataset, expanding data diversity and reinforcing its applicability in clinical scenarios with limited information.

\subsection{Results}

\begin{figure}[H]
    \centering
    \includegraphics[width=1\textwidth]{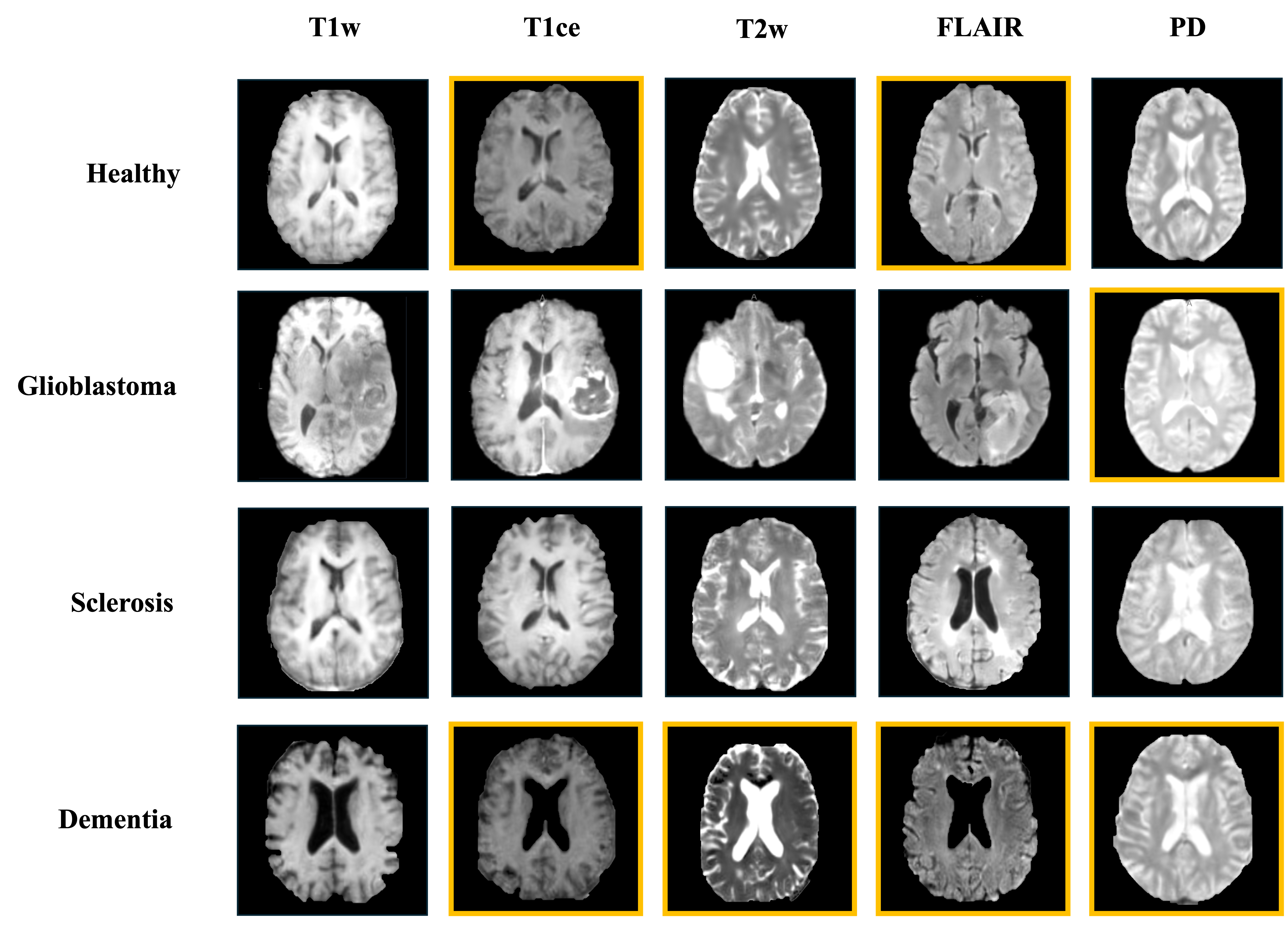} 
    \caption{Examples of images generated by the model for each combination of pathology and MRI modality. Orange borders represent data whose modality and disease were not in the training set. }
    \label{fig:imagenes_generadas}
\end{figure}

The results of MRI generation are presented in Figure~\ref{fig:imagenes_generadas}, which displays a grid of examples generated by the model. Each row represents a specific pathology (	\textit{Healthy, Glioblastoma, Sclerosis, Dementia}), while each column corresponds to an MRI modality (\textit{T1w, T1ce, T2w, Flair, PD}).

Although some pathology and modality combinations were not represented in the training dataset (highlighted in orange in Figure~\ref{fig:imagenes_generadas}), the model was able to generate 	\textbf{coherent and anatomically realistic} synthetic images, suggesting a \textbf{strong extrapolation capability}. This feature is crucial for applying the model to expand medical datasets and balance underrepresented classes.

To assess the quality of the generated images, two widely adopted metrics in generative models were used: \textbf{Fréchet Inception Distance (FID)} and 	\textbf{Multi-Scale Structural Similarity Index (MS-SSIM)} \cite{heusel2017fid, wang2003ssim}.

\begin{itemize}
    \item \textbf{Fréchet Inception Distance (FID)}: Measures the similarity between the feature distribution of generated images and real images. A lower FID value indicates that the synthetic images are more similar to the real ones \cite{heusel2017fid}. This metric has been widely used in the evaluation of generative models, including medical image synthesis \cite{kazeminia2020gans}.

    \begin{equation}
        \text{FID} = ||\mu_r - \mu_g||^2 + \text{Tr}(\Sigma_r + \Sigma_g - 2(\Sigma_r \Sigma_g)^{1/2})
    \end{equation}

        where $\mu_r$ and $\Sigma_r$ are the mean and covariance of the real images' feature representations, while $\mu_g$ and $\Sigma_g$ are the mean and covariance of the generated images' feature representations.

    \item \textbf{MS-SSIM}: Derived from the Structural Similarity Index (SSIM), it evaluates diversity within the set of generated images by analyzing the structural similarity between image pairs \cite{wang2003ssim}. It ranges between 1 and 0, where a high value indicates lower variability (more homogeneous images), while a low value suggests greater diversity \cite{zhang2018perceptual}.
    
    The SSIM score for two images $x$ and $y$ is given by:
    \begin{equation}
        SSIM(x, y) = \frac{(2 \mu_x \mu_y + C_1)(2 \sigma_{xy} + C_2)}{(\mu_x^2 + \mu_y^2 + C_1)(\sigma_x^2 + \sigma_y^2 + C_2)}
    \end{equation}
        where $\mu_x$ and $\mu_y$ are the mean intensities, $\sigma_x^2$ and $\sigma_y^2$ are the variances, and $\sigma_{xy}$ is the covariance of the images. The constants $C_1$ and $C_2$ are used to stabilize the division.

    The MS-SSIM score is computed as follows:
     \begin{equation}
        \text{MS-SSIM}(x, y) = \prod_{j=1}^{M} \left[ SSIM_j(x, y) \right]^{\alpha_j}
    \end{equation}
        where $SSIM_j(x, y)$ is the structural similarity index at scale $j$, $M$ is the number of scales, and $\alpha_j$ are the weighting factors for each scale. \cite{inproceedings}

\end{itemize}

The FID values obtained from comparing real and generated images are presented in Table~\ref{tab:fid}, showing the model's performance across each pathology and MRI modality combination.

\begin{table}[h]
    \centering
    \renewcommand{\arraystretch}{1.2}
    \begin{tabular}{|l|c|c|c|c|c|}
        \hline
        \textbf{Pathology} & \textbf{T1w} & \textbf{T1ce} & \textbf{T2w} & \textbf{FLAIR} & \textbf{PD} \\
        \hline
        Healthy & 0.005 & - & 0.001 & - & 0.007 \\
        Glioblastoma & 0.390 & 0.299 & 0.083 & 0.111 & - \\
        Sclerosis & 0.084 & 0.021 & 0.007 & 0.012 & 0.021 \\
        Dementia & 0.005 & - & - & - & - \\
        \hline
    \end{tabular}
    \caption{FID results by pathology and modality.}
    \label{tab:fid}
\end{table}

The FID results show that the model achieves high similarity between generated and real images in most modalities, especially in conditions with a larger amount of training data. Notably, lower FID values are observed in the \textit{Healthy} and \textit{Dementia} categories, indicating that generated images in these categories have a feature distribution closer to real images. In contrast, pathologies with less representation in the dataset, such as \textit{Sclerosis}, exhibit more variable FID values, suggesting a greater challenge in image synthesis for these conditions.

The \textbf{MS-SSIM} values, which assess the diversity of the generated image set, are presented in Table~\ref{tab:msssim}:

\begin{table}[h]
    \centering
    \renewcommand{\arraystretch}{1.2}
    \begin{tabular}{|l|c|c|c|c|c|}
        \hline
        \textbf{Pathology} & \textbf{T1w} & \textbf{T1ce} & \textbf{T2w} & \textbf{FLAIR} & \textbf{PD} \\
        \hline
        Healthy & 0.700 ± 0.073 & - & 0.704 ± 0.066 & - & 0.742 ± 0.059 \\
        Glioblastoma & 0.777 ± 0.051 & 0.735 ± 0.068 & 0.697 ± 0.066 & 0.673 ± 0.066 & - \\
        Sclerosis & 0.785 ± 0.055 & 0.756 ± 0.053 & 0.718 ± 0.052 & 0.766 ± 0.049 & 0.740 ± 0.069 \\
        Dementia & 0.488 ± 0.132 & - & - & - & - \\
        \hline
    \end{tabular}
    \caption{MS-SSIM results by pathology and modality.}
    \label{tab:msssim}
\end{table}

The MS-SSIM values indicate that the model generates images with a 	\textbf{good balance between similarity and structural diversity}. In particular, in \textit{Sclerosis} and \textit{Healthy} categories, the values are higher, suggesting more homogeneous images. In contrast, \textit{Dementia} exhibits the lowest value (0.488), indicating greater variability in generated images.

Overall, the MS-SSIM results show that the model is capable of generating synthetic images that maintain a 	\textbf{balance between realism and structural diversity}. However, the amount of training data influences the diversity of generated images, as categories with lower representation in the dataset exhibit greater variability in synthetic images. This suggests that a more balanced distribution of training data could enhance the model's ability to generate more diverse images across all conditions.

In summary, the FID and MS-SSIM results demonstrate that the model generates synthetic images with 	\textbf{high visual fidelity} and a \textbf{good balance between realism and diversity}. The FID values indicate that the generated images resemble real ones, particularly in categories with greater representation in the training dataset, while in less-represented pathologies, the similarity is more variable. On the other hand, the MS-SSIM values show that the model maintains sufficient diversity in generated images, avoiding excessive repetition of structures. Together, these results validate the model's ability to generate synthetic images that are both coherent and varied, even in configurations not present in the training data.

\section{Conclusions}

This work has presented a latent diffusion model (LDM) for the generation of conditioned medical images, addressing the challenges associated with the scarcity of clinical data and privacy restrictions in real image collection. Throughout the study, it has been demonstrated that the model can generate synthetic images with a high degree of realism, faithfully reproducing expected anatomical structures across different pathologies and MRI modalities.

The results obtained using the FID and MS-SSIM metrics have allowed the evaluation of both the visual fidelity of the generated images and their structural diversity. The FID values indicate that the generated images closely resemble real ones in terms of feature distribution, especially for pathologies and modalities with greater representation in the training dataset. Likewise, the MS-SSIM values show that the model maintains a balance between similarity and variability, generating diverse images without replicating exact structures from the original dataset. These results validate the model's ability to extrapolate knowledge and generate images in configurations not present in the training data, thereby expanding its applicability in medical research and the development of AI-based tools.

Additionally, the ability to generate images in any combination of pathology and modality enables its use in multiple biomedical applications. These include the expansion of clinical datasets, the creation of balanced data for training diagnostic models, the evaluation of segmentation and classification algorithms, and the simulation of clinical scenarios for trials. Medical image generation represents a viable alternative to mitigate the lack of data in rare pathologies while ensuring patient privacy without compromising the quality of trained models.

The results obtained in this work reinforce the importance of generative models in the field of AI applied to medicine. The ability of the proposed model to generate conditioned synthetic medical images opens new opportunities for biomedical research, with a significant impact on the development of diagnostic support tools and the optimization of AI systems in the healthcare sector.


\end{document}